\documentclass[letterpaper, 10 pt, conference]{ieeeconf}  

\IEEEoverridecommandlockouts                              

\usepackage[pdftex]{graphicx} 
\usepackage{amsmath,amssymb}
\usepackage{bm}
\usepackage{subfig}

\title{\LARGE \bf
Learning 6DoF Grasping Using Reward-Consistent Demonstration
}

\author{Daichi Kawakami$^{1}$, Ryoichi Ishikawa$^{1}$, Menandro Roxas$^{2}$, Yoshihiro Sato$^{3}$, Takeshi Oishi$^{1}$
\thanks{$^{1}$Daichi Kawakami, Ryoichi Ishikawa and Takeshi Oishi are with Institute of Industrial Science, The University of Tokyo, Japan {\tt\small \{dkawakami, ishikawa, oishi\}@cvl.iis.u-tokyo.ac.jp}}%
\thanks{$^{2}$Menandro Roxas is with LINE Corporation {\tt\small menandro.roxas@linecorp.com}
}
\thanks{$^{3}$Yoshihiro Sato is with Faculty of Engineering, Kyoto University of Advanced Science, Japan {\tt\small sato.yoshihiro@kuas.ac.jp}
}
}

\begin{document}

\maketitle
\thispagestyle{empty}
\pagestyle{empty}

\begin{abstract}

As the number of the robot's degrees of freedom increases, the implementation of robot motion becomes more complex and difficult. 
In this study, we focus on learning 6DOF-grasping motion and consider dividing the grasping motion into multiple tasks. We propose to combine imitation and reinforcement learning in order to facilitate a more efficient learning of the desired motion.
In order to collect demonstration data as teacher data for the imitation learning, we created a virtual reality (VR) interface that allows humans to operate the robot intuitively. Moreover, by dividing the motion into simpler tasks, we simplify the design of reward functions for reinforcement learning and show in our experiments a reduction in the steps required to learn the grasping motion.

\end{abstract}
\section{Introduction}

Development of humanoid robots with high degrees of freedom (DOF) and can realize complex motions is becoming more popular. 
As the DOF increases, more advanced intelligence becomes necessary to control these robots.

When robots perform tasks (e.g. serving, cleaning, medical care, disaster rescue, etc.), grasping an object becomes an essential motion.
The mere variety of the type of robot hand they have and the need for handling new objects in different situations opens up a plethora of intractable grasping motions. Manually implementing these actions for every situation is impossible, therefore, it becomes more important to have an automated learning of the grasping motion. 

Compared to humans, robots can only gather and utilize limited information. 
For example, using cameras attached to the robot usually suffers from occlusion problems especially during handling of an object. 
Humans compensate with tactile information but in many cases, robots do not have tactile sensors.
One approach to address this is to estimate the mass and center of gravity of an object using RGB images \cite{estimatemass}, but for common transparent objects e.g. water bottles, this technique fails.
Another approach is estimating the frictional force, but since it changes depending on the surface condition of the object, relying on it is impractical. Additionally, handling of an object changes depending on its usage.
In existing works, robots mainly utilize the pose  of an object from RGB-D information to decide the handling approach.
Needless to say, the discrepancy in information makes it difficult for robots to achieve the same grasping motion as humans. 

Even though datasets for learning grasp point detection exist \cite{cornel}, 
we still need to abstract them to be usable for every hardware. 
Using these datasets, realizing the grasping motion for different hardware, such as complex multi-fingered hands or soft fingers, becomes a daunting task especially because we need to consider other factors beside just the grasping point and hand angle \cite{dex, soft}.


Reinforcement learning is an effective method for robots to learn grasping motions when ground truth datasets are insufficient.
However, for complex motions, the design of rewards is slow especially because one trial in which the robot needs to move takes time and accumulation of sufficient number of trials can take days.

To address this issue, imitation learning, where in a robot learns from a human demonstration of its own operation, were proposed  \cite{imitate1, imitate2}.
Many human-robot interfaces for this domain have been proposed, such as master-slave systems \cite{vinci}.
However, these interfaces usually require expensive equipment on the operator's side specific to the associated robot, which makes master-slave systems impractical to be applied on a wide variety of robots. Considering that different robot designs will be available in the future, a less expensive operation interface is obviously necessary.

\begin{figure}[tbp]
    \centering
    \includegraphics[width=8cm]{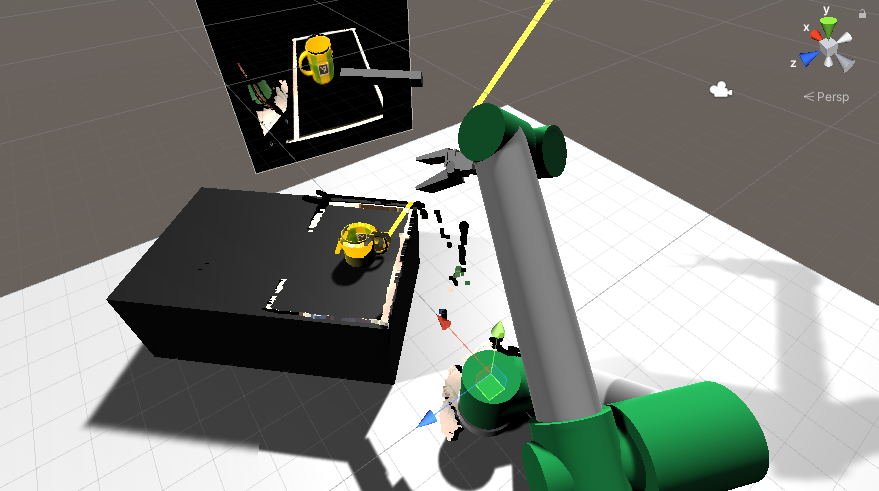}
    \caption{VR interface for robot control}
    \label{fig_vr_interface} 
\end{figure}

In this work, we propose a framework to gracefully combine reinforcement learning and imitation learning with a Virtual Reality (VR) interface.
Generally, the reward design of reinforcement learning and demonstration of imitation learning might be inconsistent, in which case, the training of reinforcement learning will never converge.
Hence, the proposed method combines the learning processes by dividing the motion into multiple tasks.
We also develop a robot operation interface in VR space, where the operator can intuitively control the robot while recognizing 3D information in VR, as shown in Fig. \ref{fig_vr_interface}.
We aim to collect the demonstration data of 6-DOF grasping using this interface.


\section{Related work}

Most of the research in robot grasping deals with 3-DOF grasping in which 2D points and 1D angles are given from one perspective of the handled object \cite{2dsota, dexnet}. 

In this domain, deep learning methods for grasp point detection have been proposed \cite{deep}.
This detection for 3-DOF grasping has since greatly improved \cite{pointnetgpd} due to efficient point cloud processing techniques \cite{pointnet}.

In \cite{graspinginteraction}, dataset creation for 6-DOF grasp point detection was proposed by annotating the points in VR space. In some works \cite{vr1, vr2, vr3}, VR is also used in performing robot operations.
On the other hand, Mousavian et al. \cite{learning6dof} attempted to estimate the grasp point using physics simulations in \cite{flex} and 3D models of objects. Although using simulation data removes the need for human annotation, more computational resources is necessary to achieve a more realistic operation. Additionally, the differences between real world and simulations cannot be avoided.

Some works \cite{google, supersizing} use reinforcement learning to learn robot grasp. Trials were conducted for tens of days in which the robot needed to actually grasp an object which makes data collection time consuming.
In addition, reinforcement learning without human-annotated data may result in learning non-human-like behaviors \cite{nothuman}.

In contrast, imitation learning
can provide the mapping between robot states and motions in a shorter time \cite{fish}.
However, the demonstration data required to learn to grasp various objects tends to be large.
In existing works, the demonstration data required for grasping one type of object needs about 10 minutes of demonstration data \cite{imitation}.
Imitation learning also incorporates behavioral cloning \cite{bc}, which simply maps states to actions, and inverse reinforcement learning \cite{irl}, which estimates rewards from demonstration data. Generative adversarial imitation learning (GAIL) \cite{gail} were also used, which is a more efficient imitation learning method.

 
\section{Motion Restructuring}

In this section, we address the complexity of designing reward functions for reinforcement learning, especially for a series of complex motions. Our approach includes dividing the target motion into multiple tasks, which we will describe in detail below.

\begin{figure}[tbp]
    \centering
    \includegraphics[width=8cm]{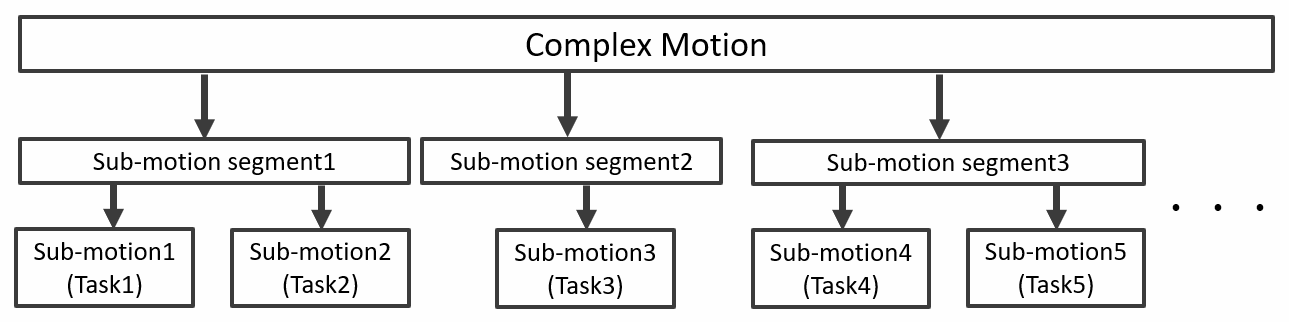}
    \caption{Motion division into multiple tasks}
    \label{fig-task-division} 
 \end{figure}

\subsection{Sub-motion Definition}
First, the target motion ${\Omega}$ is divided into semantic sub-motion segments ${\Omega^{sem}_i}$ ($i\in \mathbb{N}$) considering a motion sequence as shown in Fig. \ref{fig-task-division}. 
For example, the grasping motion can be divided into two motions: the robot hand approaching the target and grabbing.
The semantic sub-motion segments are also divided into sub-motions ${\Omega^{task}_j}$ ($j\in \mathbb{N}$) through the geometric constraints. 
\begin{equation}
    \Omega = \Omega^{sem}_1 \cdot \Omega^{sem}_2 \cdots = (\Omega^{task}_1 \cdot \Omega^{task}_2) \cdot (\Omega^{task}_3) \cdots .
\end{equation}
For example, the robot hand's approaching of a target can be divided into a movement to face the target and a movement to reduce the distance to the target.

\subsection{Global Optimization of Multiple Task Framework}

\begin{figure}[tbp]
    \centering
    \includegraphics[width=7cm]{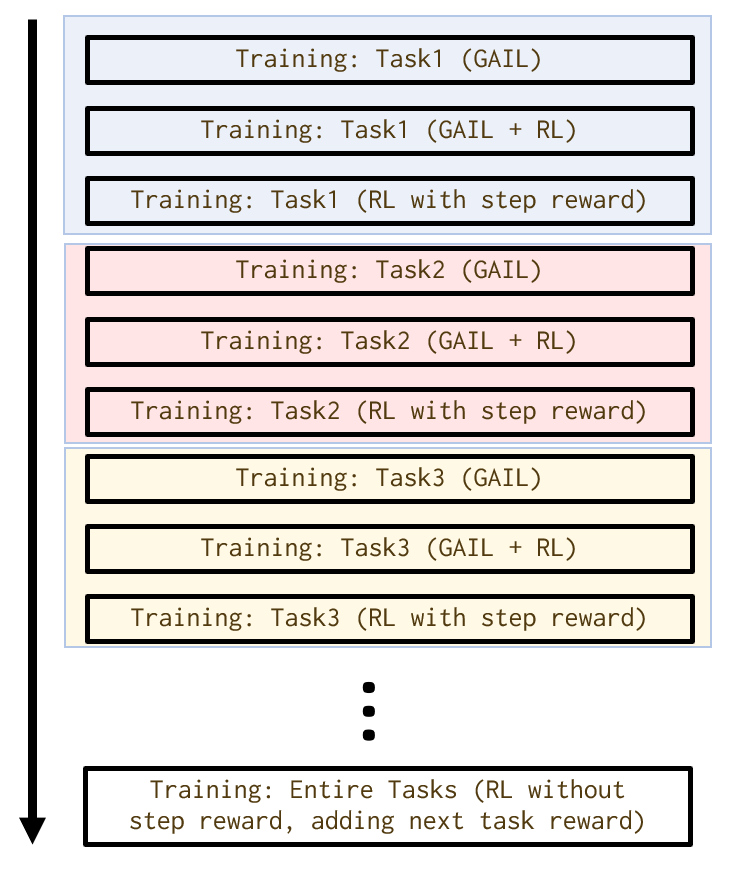}
    \caption{Learning flow for divided tasks}
    \label{fig-task-cascade} 
 \end{figure}
 
We consider constructing a neural network for individual tasks.
For each task, we first pretrain with GAIL, decrease the learning rate of GAIL and conduct reinforcement learning to get robust motion as shown in Fig. \ref{fig-task-cascade}.
After learning the motion of the task, we then conduct a reinforcement learning only optimization.

Let $s_t$ be the state of the environment at step $t$.
$r_t$ is the reward at step $t$.

Value function $V(s_t)$ with state $s_t$ is defined as follows, using discount rate $\gamma$.
\begin{equation}
    V(s_t) = \displaystyle E \left[ \sum_{k = 0}^\infty \gamma^{t+k}r_{t+k} \right]
\end{equation}

Generalized Advantage Estimation (GAE) \cite{gae} is expressed as follows, using discount rate $\lambda$.
\begin{equation}
    A(s_t) =  \sum_{k = 0}^\infty (\lambda \gamma)^k\delta_{t+k}^{\gamma}
\end{equation}
where
\begin{equation}
    \delta_{t} = r_t + \gamma V(s_{t+1}) - V(s_t)
\end{equation}

In reinforcement learning, a small negative reward $r_{step}$ can be given to each step to optimize and reduce the total number of steps required to achieve the desired behavior. 
When the step reward $r_{step}$ is added, the value function $V^{step}(s_t)$ is expressed as follows.
\begin{equation}
    V^{step}(s_t) = V(s_t) + r_{step} \displaystyle E \left[ \sum_{k = 0}^\infty \gamma^{t+k} \right]
\end{equation}
To minimize $\displaystyle E \left[ \sum_{k = 0}^\infty \gamma^{t+k} \right]$, the number of total steps of an episode should be reduced.
Therefore, the motion can be optimized by adding step reward $r_{step}$.

However, even though each individual task is optimized, we cannot assume the same when it is combined with other tasks.
After learning the individual tasks using separate neural networks, we have to consider optimizing the whole motion.
We do this by adding the sum of the rewards $U(\Omega_{i+1}^{task})$ of each task $\Omega^{task}_i$ at the completion of the next task $\Omega^{task}_{i+1}$.
\begin{equation}
    U(i) = \sum_{\Omega_{i}^{task}} r_t
\end{equation}
Let $t_{end}^{i}$ be the completion step of $\Omega_{i}^{task}$.
Next task reward $r_{next\_task}^i$ is added at $t_{end}^{i}$ and expressed as follows.
\begin{equation}
    r_{next\_task}^i = U(i+1)
\end{equation}
When the next task reward $r_{next\_task}^i$ is added, value function $V^{next\_task}$ is expressed as follows.
\begin{equation}
\begin{split}
    V^{next\_task}(s_t) = V(s_t) +\gamma^{t_{end}^{i+1} - t}&r_{next\_task}^i \\
    & (t_{end}^{i-1} \leq t < t_{end}^i)
\end{split}
\end{equation}
Since $\gamma < 1.0$, $t_{end}^i$ should be reduced to maximize $V^{next\_task}(s_t)$.
Therefore, the motion can be optimized by adding next task reward $r_{next\_task}^i$.

\subsection{Adaptive Reward Design}

Dividing the motion may ease the rewards design for reinforcement learning.
However, it is still possible that multiple rewards may conflict with each other and interfere with the learning of the target behavior. 
In order to learn the target behavior, it is effective to change the value of the reward in steps according to the learning progress.
To realize the target motion, the conditions with the highest priority needs to be learned first, and as such, by setting a large reward value for high-priority conditions in the initial stage of learning, these conditions can be learned even if the learning of the low-priority conditions is insufficient.
After some steps of the learning process, if the high-priority conditions are achieved, the reward value for the low-priority conditions can be increased to enable the learning of both. 
\section{Demonstration in VR Space}

In this section, we describe how humans can operate robots in VR space and collect demonstration data as the robots perform grasping motions.

\subsection{VR Interface for Robot Operation}

During grasping, the hand and the object are in contact with each other and therefore
requires real-time operation of the robot.
In this work, we use a motion mapping-based VR interface in which humans can operate robots intuitively using inexpensive equipment.

In VR space, we create a prior 3D model of the robot. During operation, the joint angles of this robot are set according to the their real world counterparts.
The target objects are then represented in VR space using the pose estimated by visual information.
By displaying the 3D model and the point cloud at the same time in the workspace, the confidence of the pose estimation becomes apparent.




\subsection{Robot Operation by Continuous Inverse Kinematics}
\label{sec:robotoperation}

When a human manipulates a robot, the position $p \in \mathbb{R}^3$ and the posture $\phi \in \mathbb{R}^3$ of the robot hand in 3D space must be determined according to the human input.
Let $p_{target} \in \mathbb{R}^3$ be the target position of the robot hand and $\phi_{target} \in \mathbb{R}^3$ be the target posture.
Inverse kinematics is used to determine the joint angles of the manipulator and change the current position and posture of the robot hand from $p$ and $\phi$ to the target $p_{target}$ and $\phi_{target}$.
If the dimension of the target point is six dimensions and the dimension of the robot arm is more than six dimensions, the angles of the joints can be obtained analytically, but a small change in the target point may result in a large difference in the angles of the joints that realize the target point.
To solve this problem, we use Jacobian inverse kinematics.
Let $\theta$ be the angle of each joint of the robot, and let $\Delta q$ be the small change in $p$ and  $\phi$ when $\theta$ changes by a small amount.
$\Delta$q can be expressed using the Jacobian $J$.

\begin{equation}
    J = \frac{\partial \bm{q}}{\partial \bm{\theta}}
    \end{equation}
    \begin{equation}
    \Delta {\bm{q}} = 
    \begin{bmatrix}
    \Delta \bm{p} \\
    \Delta \bm{\phi} \\
    \end{bmatrix}
    = J
    \Delta \bm{\theta}
\end{equation}

Then, the direction of movement of each joint $\Delta \theta$ can be expressed using the generalized inverse matrix $J^{\sharp}$ of $J$.

\begin{equation}
    \Delta {\bm{\theta}} = J^{\sharp} \Delta {\bm{q}}
\end{equation}

The $\Delta \theta$ calculated in this way is repeatedly added to the current joint angle of the arm to bring the robot hand to the target point.
If the $\Delta q$ is large, there is a possibility that each joint of the robot will move significantly.
In this case, it may make the intended human manipulation difficult.
Therefore, it is necessary to apply constraints to prevent $ \Delta {\bm{q}}$ from becoming too large.

The opening and closing motions of the robot hand are given different inputs from the target position and orientation of the robot hand.
The input methods include key input by the operator and tracking of the operator's hand to calculate the distance between the tips of index finger and thumb.

\section{Implementation}

\subsection{Hardware}

\begin{figure}[tbp]
    \centering
    \includegraphics[width=6cm]{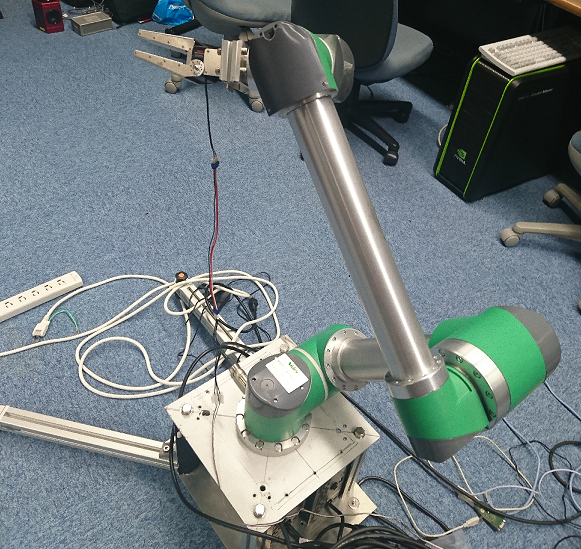}
    \caption{7-DOF manipulator, Arm: 6-DOF, Hand: 1-DOF}
    \label{fig_robot} 
\end{figure}

The robot arm and hand are shown in Fig. \ref{fig_robot}. For the robot arm, we use a 6-DOF manipulator \footnote{Nidec Corporation, i611.} and for the robot hand, we use a hand-made parallel two-finger gripper.
Realsense SR305\footnote{ https://www.intel.com/content/www/us/en/architectureand-technology/realsense-overview.html} is used as the RGBD sensor.

\subsection{Grasping Motion Reconfiguration}

We divide the grasping motion into the following three tasks.
\begin{itemize}
\item Task 1: Facing the robot hand toward the predetermined grasping direction.
\item Task 2: Approaching the robot hand to the grasping point while satisfying the constraints of Task 1.
\item Task 3 : Closing the robot hand and grasping the object.
\end{itemize}

\begin{figure}[tbp]
    \centering
    \includegraphics[width=6cm]{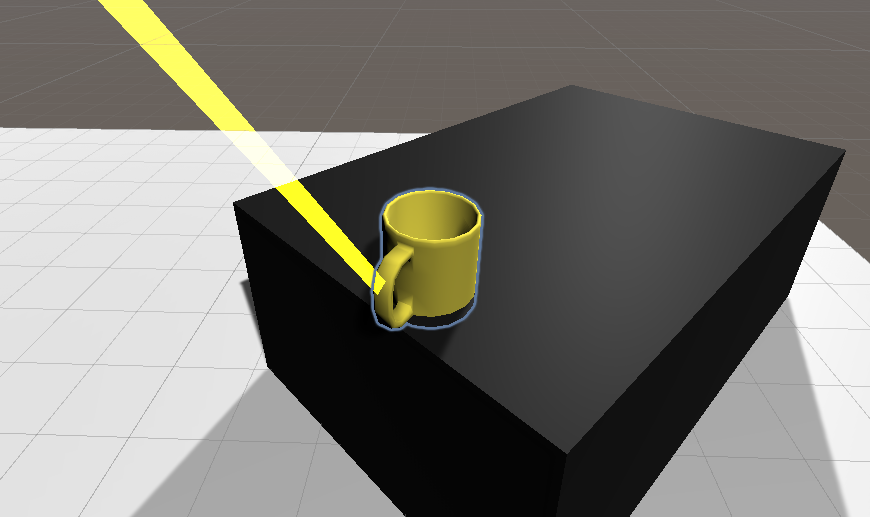}
    \caption{Grasping direction}
    \label{fig_grasping_direction} 
\end{figure}

The grasping direction is determined for the target object as shown in Fig. \ref{fig_grasping_direction}.
In this experiment, a yellow cup with a handle is used.
Multiple grasping directions can be determined, but only one is used.


\subsection{Reward Design}


In addition to changing the value of the reward according to the learning progress, we also perform curriculum learning in which the type of reward is changed.
For each of the divided tasks, imitation learning is performed in GAIL first.
When the success rate of each task exceeds a certain value, learning method shifts to reinforcement learning for each task.
When the reinforcement learning of Task 1 is completed, the learning of Task 1 is stopped and the learning of Task 2 is started.
When learning Task 2, the initial state of Task 2 is generated using the learning results of Task 1.
Similarly, when learning Task 3, the initial state is generated using the learning results of Task 1 and Task 2. When the reinforcement learning of Task 3 is completed, we move on to the optimization of the behavior of the entire motion.

In order to compare with the pure imitation learning result and pure reinforcement learning without motion division, two more experiments are conducted.
First, we perform GAIL only from obtained demonstration.
Second, only reinforcement learning with the rewards that is designed by combining rewards for Task 1 to Task 3 is conducted.

\subsubsection{Task 1}

The timing of feeding reward in Task 1 is designed as follows.
\begin{itemize}
\item When the direction of the robot hand approaches the grasping direction
\item When the position of the robot hand approaches the grasping direction
\item When the robot hand reaches the grasping direction
\item When a certain number of steps have elapsed
\item When the robot collides with the target object or the environment
\item When the robot hand moves away from the object by more than a certain distance
\end{itemize}
In addition, we change the reward according to the learning progress as follows.
\begin{itemize}
\item When the success probability of Task 1 exceeds a certain value.

In order to optimize the motion, a small negative reward is given as a punishment for each step.
\item When the learning of all tasks is completed

To optimize the motion throughout the tasks, giving step reward is stopped. When Task 2 is completed, the sum of the rewards for Task 2 is given.
\end{itemize}

\subsubsection{Task 2}
The timing of feeding reward in Task 2 was designed as follows.
\begin{itemize}
\item When the robot hand approaches the grasping point
\item When the robot hand is not facing the grasping direction
\item When a certain number of steps have elapsed
\item When the robot collides with the target object or the environment
\item When the robot hand moves away from the object by more than a certain distance
\end{itemize}
In Task 2, the learning process requires the robot hand to avoid colliding with the target object. Considering this, the reward is changed according to the learning progress as follows.
\begin{itemize}
\item When a certain number of steps have elapsed

The reward when the robot hand collides with an object is increased. The robot may have learned to avoid colliding with the target object without reaching the target through the reward at the time of collision. 
This is  to reduce the impact of the punishment at the time of collision.
\item When Task 2 succeeds

In Task 2, a reward (punishment) for colliding with an object is given. This reward is decreased when Task 2 succeeds which increases the impact of the punishment for the collision.
\item When the probability of success in Task 2 exceeds a certain value

In order to optimize the motion, a small negative reward is given as a punishment for each step.
\item When the learning of all tasks is completed

To optimize the motion throughout the tasks, giving step reward is stopped. When Task 2 is completed, the sum of the rewards for Task 2 is given.
\end{itemize}
\subsubsection{Task 3}
The timing of feeding reward in Task 3 was designed as follows.
\begin{itemize}
\item When the hand closed to a certain angle at grasping point
\item When a certain number of steps have elapsed
\item When the hand collided with the environment
\item When the robot hand moves away from the object by a certain distance
\end{itemize}

\subsection{Interface}
In order to get demonstration data for imitation learning, we implemented VR interface for robot operation.
We use the Oculus Quest\footnote{https://www.oculus.com/quest/ for this study.
}.

\begin{figure*}[t]
    \centering
    \includegraphics[width=17cm]{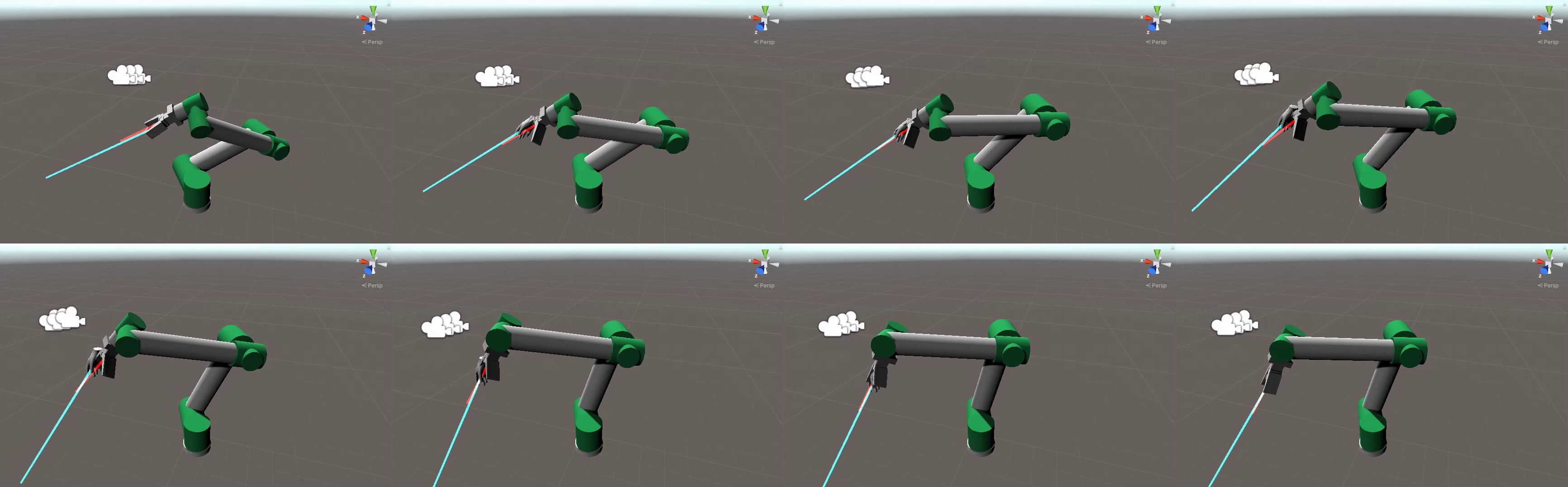}
    \caption{Operation interface. Red line: hand direction, Blue line: target direction}
    \label{fig_ik}
\end{figure*}

\begin{figure*}[!t]
    \centerline{
    \subfloat[Task 1]{\includegraphics[clip, width=5cm]{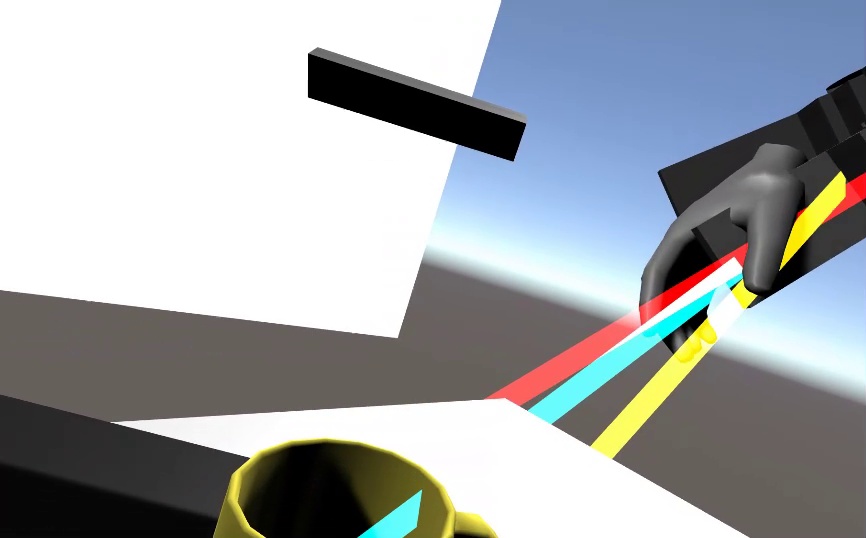}
    \label{fig:label-A}}
    \hfil
    \subfloat[Task 2]{\includegraphics[clip, width=5.0cm]{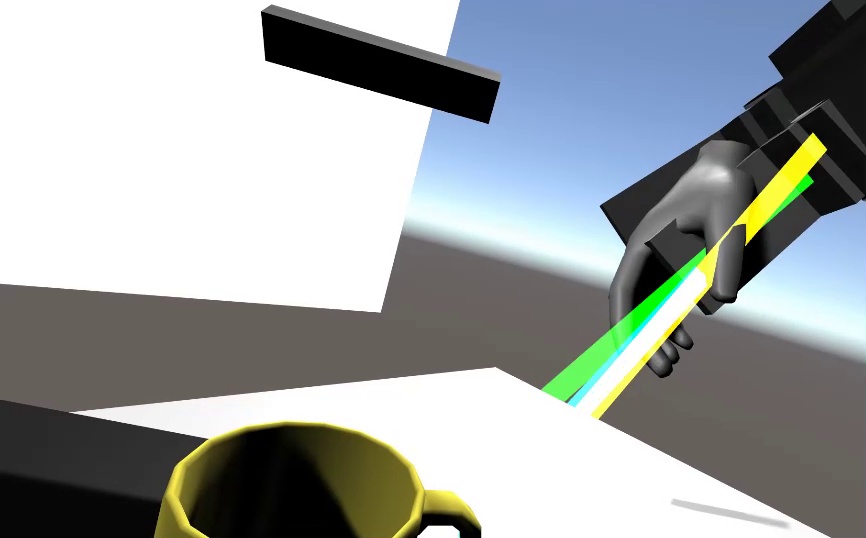}
    \label{fig:label-B}}
    \hfil
    \subfloat[Task 3]{\includegraphics[clip, width=5.0cm]{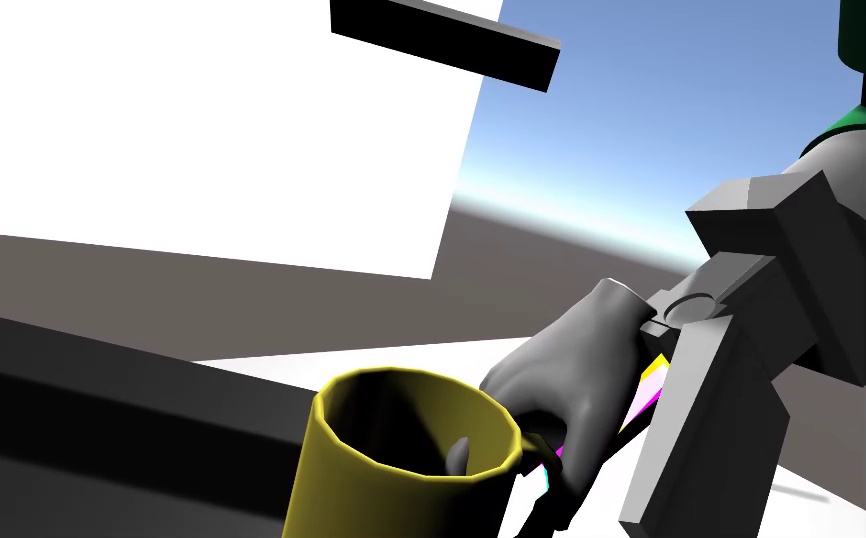}
    \label{fig:label-C}}
    }
    \caption{Task state indication during demonstration}
    \label{fig_task_state} 
\end{figure*}

An Operator controls the robot using a controller device with tracked position and pose (i.e. Oculus Quest Controller) which then translates to the target position and pose of the robot hand.
The angle of each joint to be moved is calculated as described in Section \ref{sec:robotoperation}. 
During the operation, the task state is indicated by the color of the line pertaining to the direction of the robot hand as shown in Fig. \ref{fig_task_state}.

As input features, the 3D position and rotation (in quaternion) from joints of arm, joints of hand, end-effector, and target are used, with total dimension of $7 \times (6 + 1 + 1 + 1) = 63$.
As output features, the angular velocities of robot joints are used. The total dimension is $7$.
Using the implemented interface, demonstration data of 50 episodes were obtained in the simulation environment.

\subsection{DNN Architecture and Learning Algorithm}

\begin{figure}[tbp]
    \centering
    \includegraphics[width=8cm]{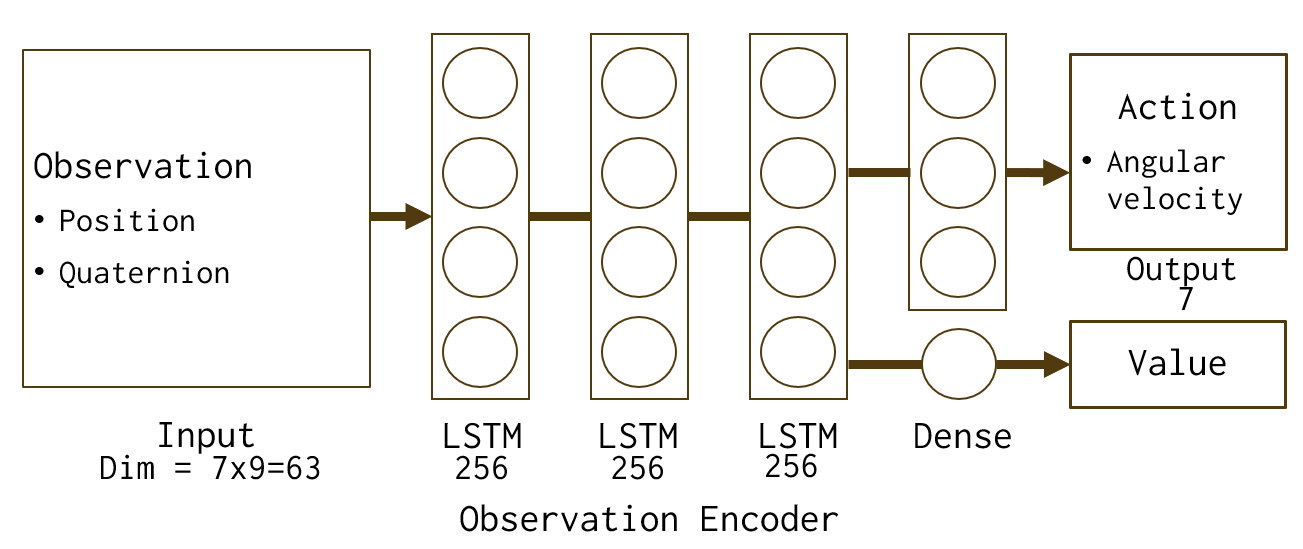}
    \caption{DNN architecture for Task 1 - Task 3}
    \label{fig_dnn} 
\end{figure}

The DNN models to be trained in each task are separated.
Fig. \ref{fig_dnn} shows the network architecture used in Tasks 1 to 3.
The input layer is followed by 3 LSTM \cite{lstm} layers with 256 nodes in each hidden layer.
In order to generate the output features and the estimated value function, fully connected layer (dense layer) follows the LSTM layers.
We use the PPO (Proximal Policy Optimization) as the learning algorithm \cite{ppo}.


\section{Experiment}


\subsection{Imitation Learning}

\begin{figure}[t]
    \centering
    \includegraphics[width=8cm]{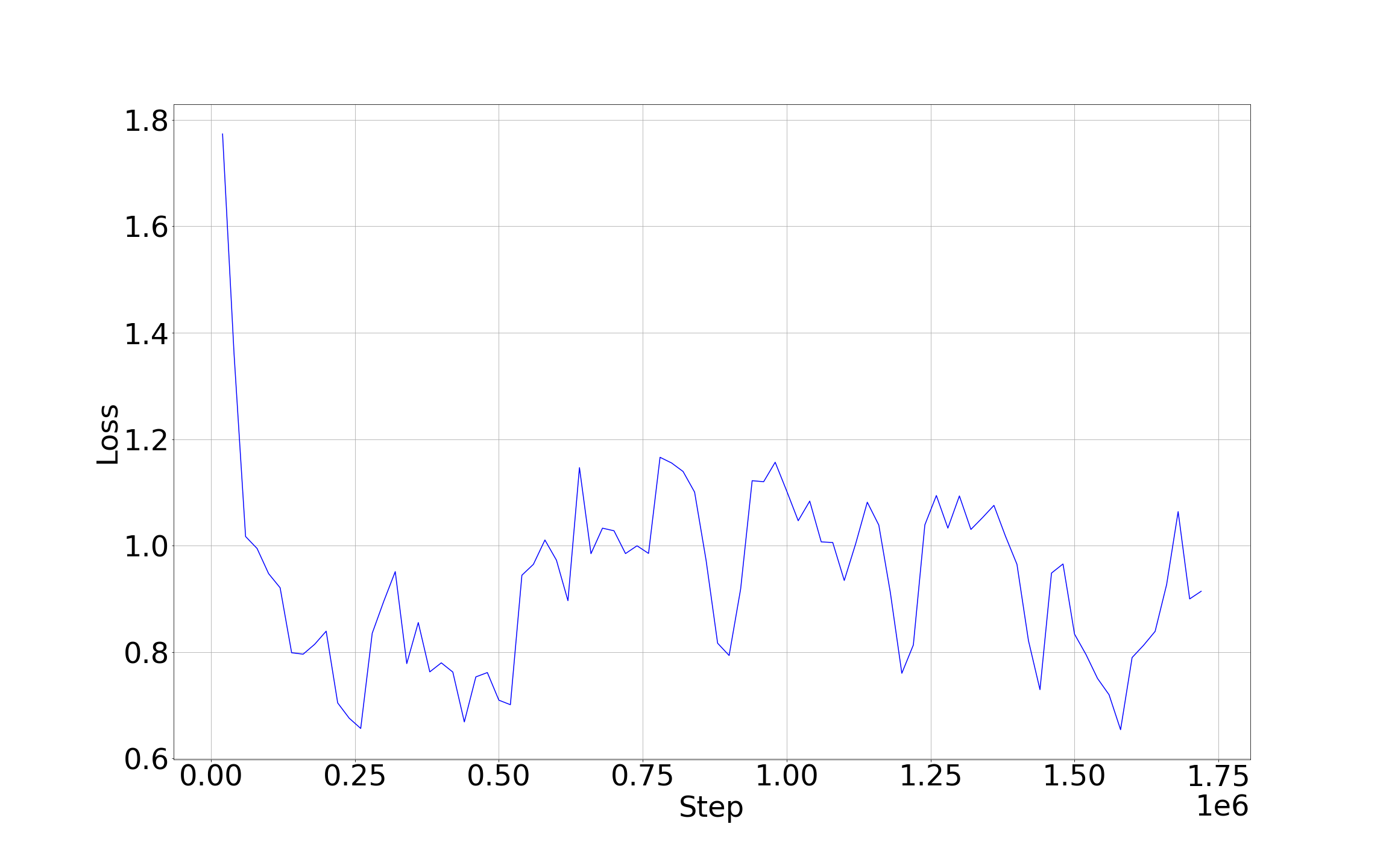}
    \caption{GAIL Loss. Grasping motion is not divided into tasks.}
    \label{task_all_gail_loss}
\end{figure}

\begin{figure}[t]
    \centering
    \includegraphics[width=8cm]{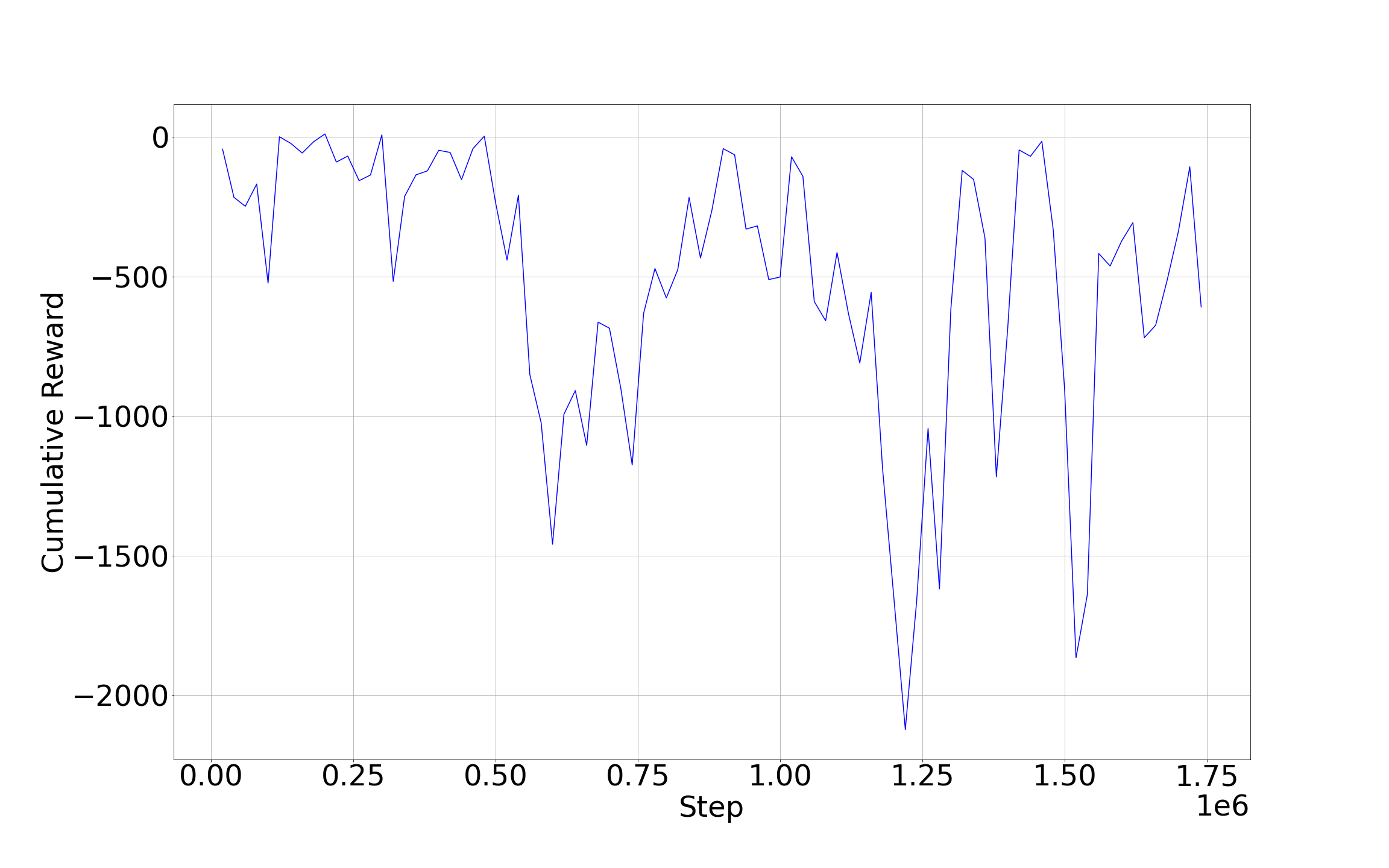}
    \caption{Average cumulative reward per 1 episode on GAIL. Grasping motion is not divided into tasks. This reward is not used for learning.}
    \label{task_all_gail_reward}
\end{figure}

For the obtained demonstration, we only perform GAIL.
Figs. \ref{task_all_gail_loss} and \ref{task_all_gail_reward} show GAIL loss and cumulative reward per 1 episode for every 20,000 steps.
In this experiment, the value of grasping success reward is 100.0.
The GAIL loss is reduced in the early steps (100,000 steps). However, the cumulative reward never exceeds the value of grasping success reward.
This is because the demonstration data is not sufficient for various initial state of the environment.

\subsection{Reinforcement Learning}

\begin{figure}[t]
    \centering
    \includegraphics[width=8cm]{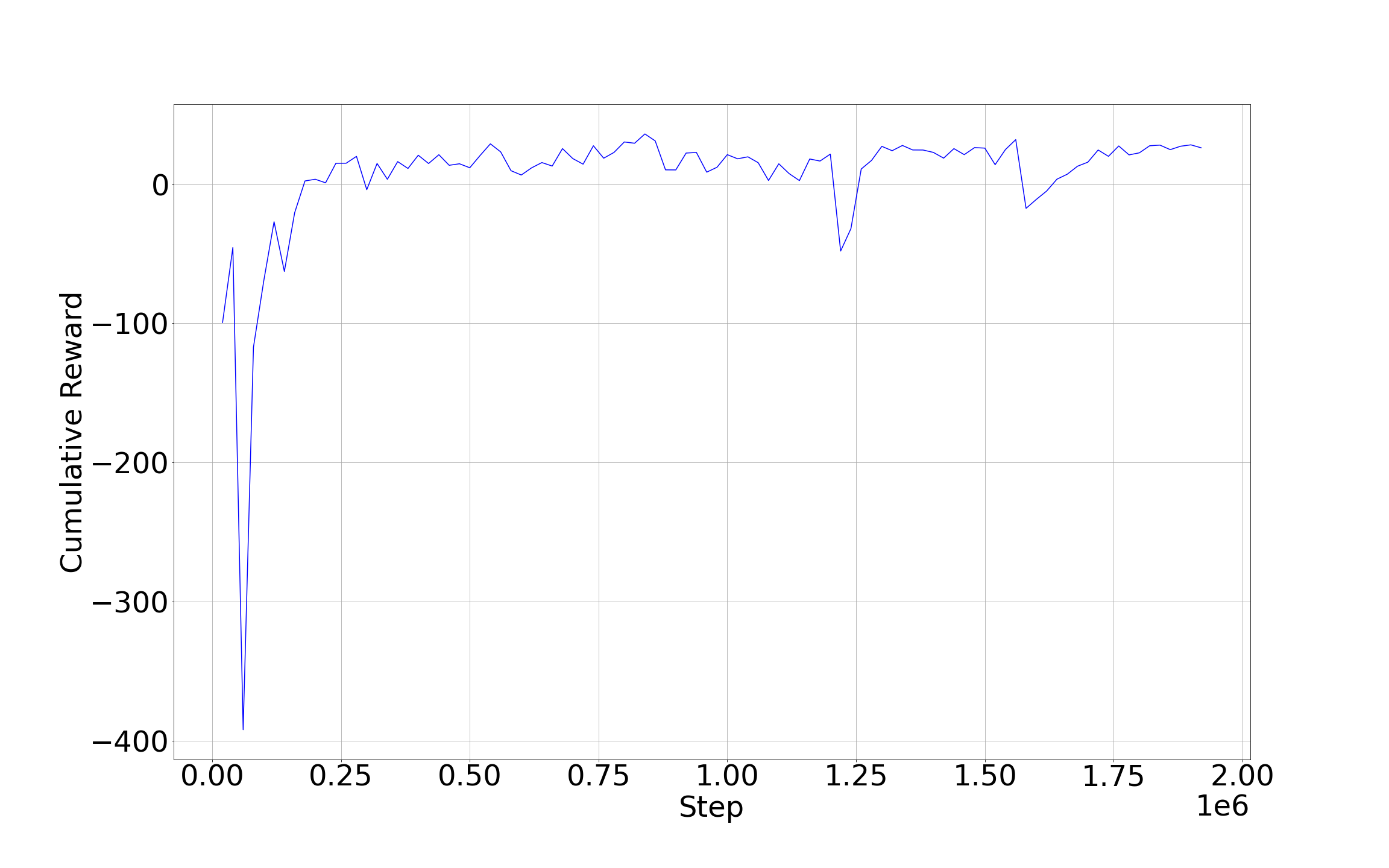}
    \caption{Average cumulative reward per 1 episode. Grasping motion is not divided into tasks. Learning starts from reinforcement learning.}
    \label{task_all_reward}
\end{figure}

We then conduct reinforcement learning with rewards designed through combining the rewards for Tasks 1 to 3.
Fig. \ref{task_all_reward} shows cumulative reward per 1 episode for each 20,000 steps.
After 250,000 steps, the cumulative reward stopped increasing and is less than the grasping success reward.
In the learning process, the robot first learn to approach the target.
At first, the robot hand easily collided with the target and the robot starts to learn collision avoidance.
However, the robot tends to collide with the environment and never reaches the grasping point without collision.
At some point, the reward becomes a local optimum and the robot never learn the grasping motion.
In this experiment, the robot could not learn closing the hand at the grasping point.

\subsection{Fusion of Imitation Learning and Reinforcement Learning}

\begin{figure}[tbp]
    \centering
    \includegraphics[width=7cm]{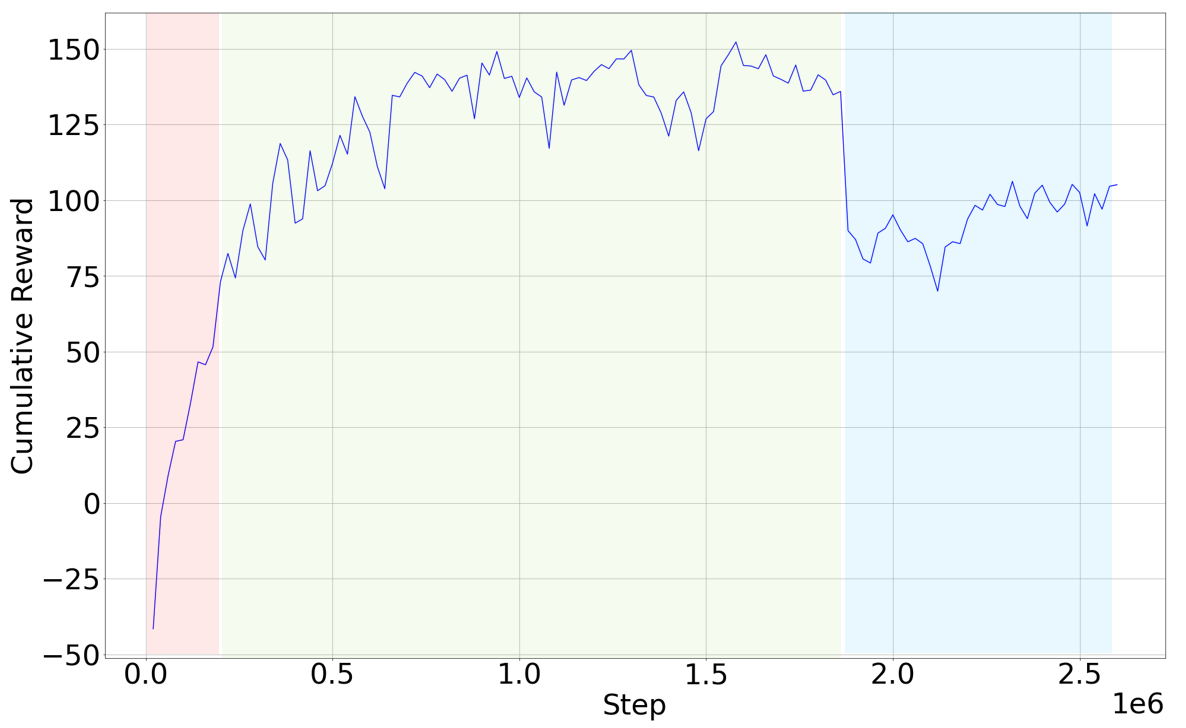}
    \caption{Average cumulative reward per 1 episode on Task 1. Red: GAIL, Green: GAIL + RL, Blue: RL + step reward}
    \label{task1-reward}
\end{figure}

\begin{figure}[tbp]
    \centering
    \includegraphics[width=7cm]{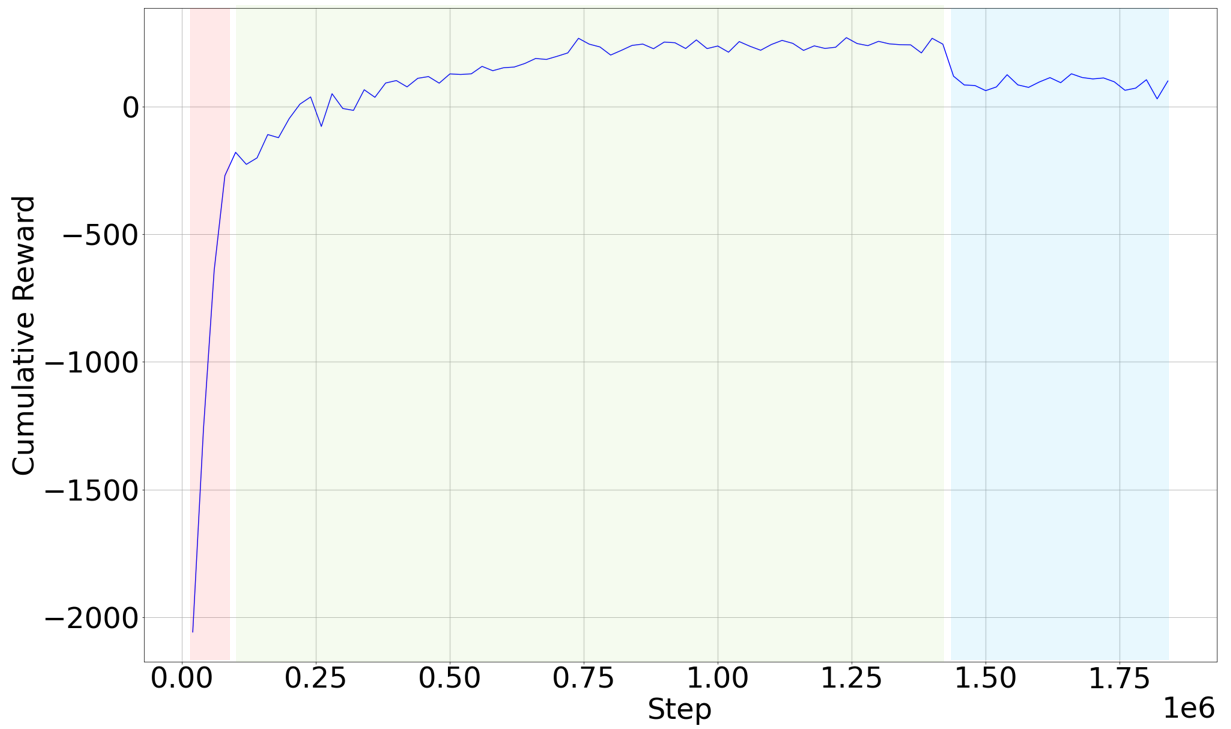}
    \caption{Average cumulative reward per 1 episode on Task 2. Red: GAIL, Green: GAIL + RL, Blue: RL + step reward}
    \label{task2-reward}
\end{figure}

\begin{figure}[tbp]
    \centering
    \includegraphics[width=7cm]{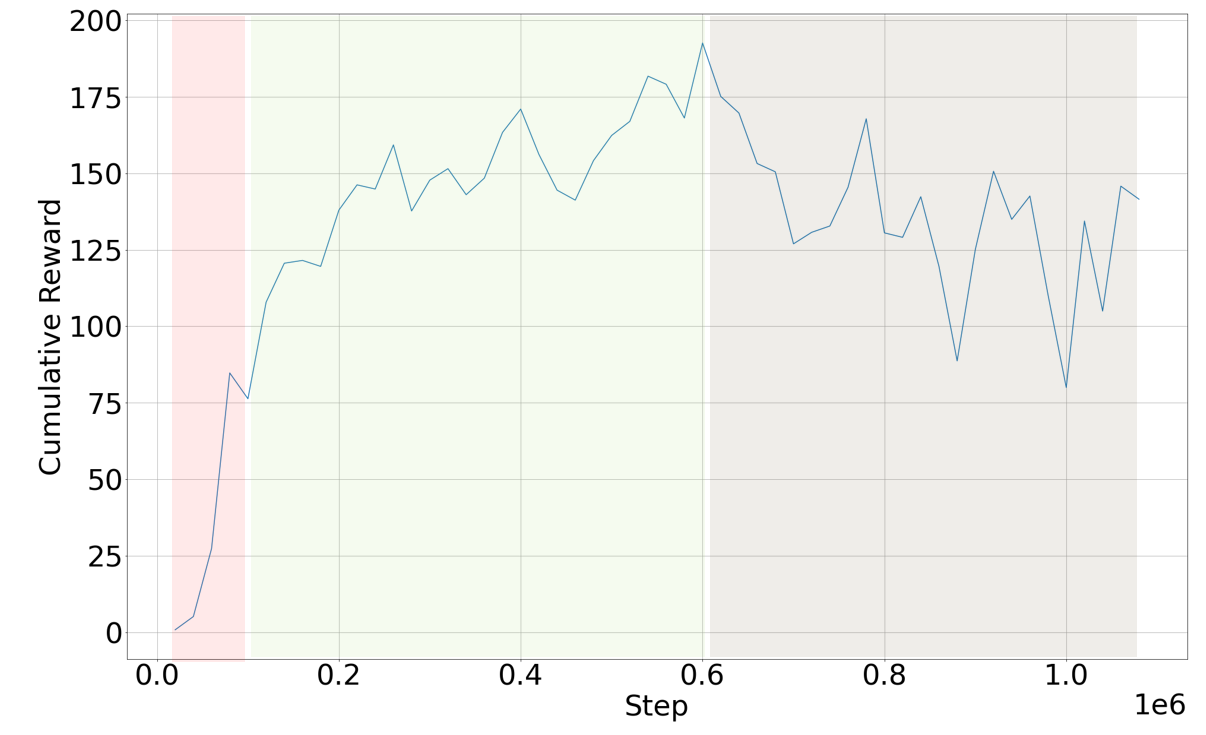}
    \caption{Average cumulative reward per 1 episode on Task 3. Red: GAIL, Green: GAIL + RL, Gray: Optimization for entire tasks}
    \label{task3-reward}
\end{figure}

\begin{figure}[tbp]
    \centering
    \includegraphics[width=7cm]{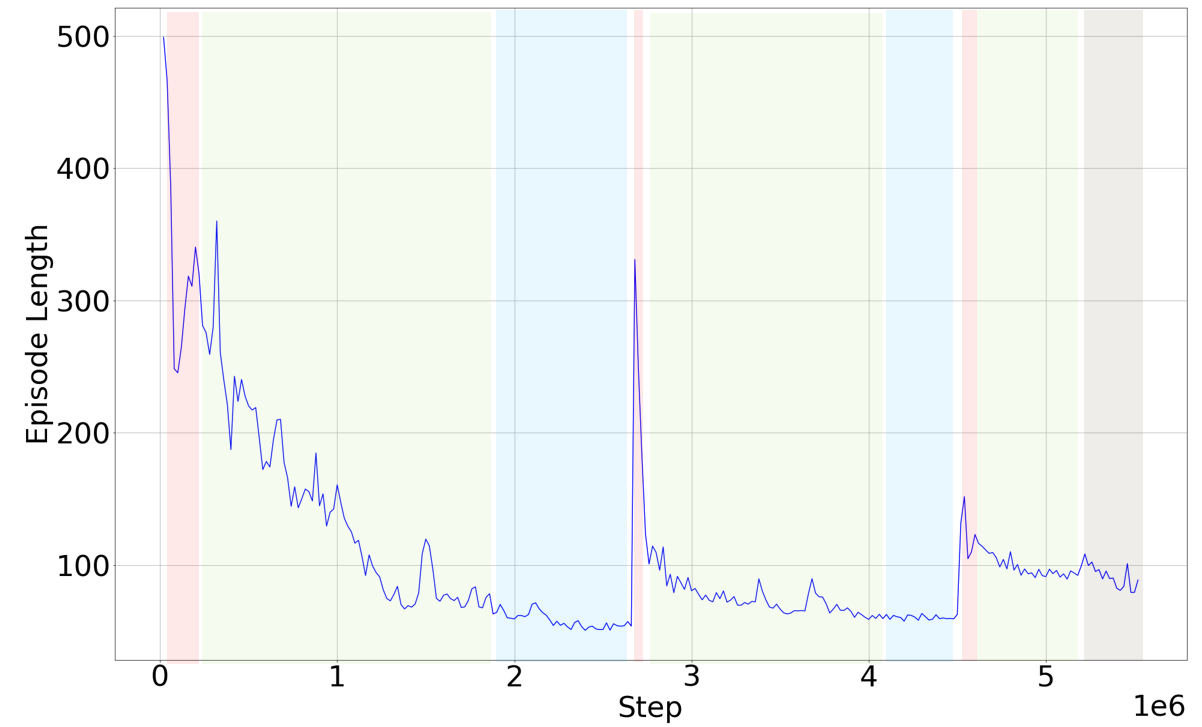}
    \caption{Average number of steps per 1 episode. Red: GAIL, Green: GAIL + RL, Blue: RL + step reward, Gray: Optimization for entire tasks}
    \label{tasks-step}
\end{figure}

\begin{figure*}[tbp]
    \centering
    \includegraphics[width=16cm]{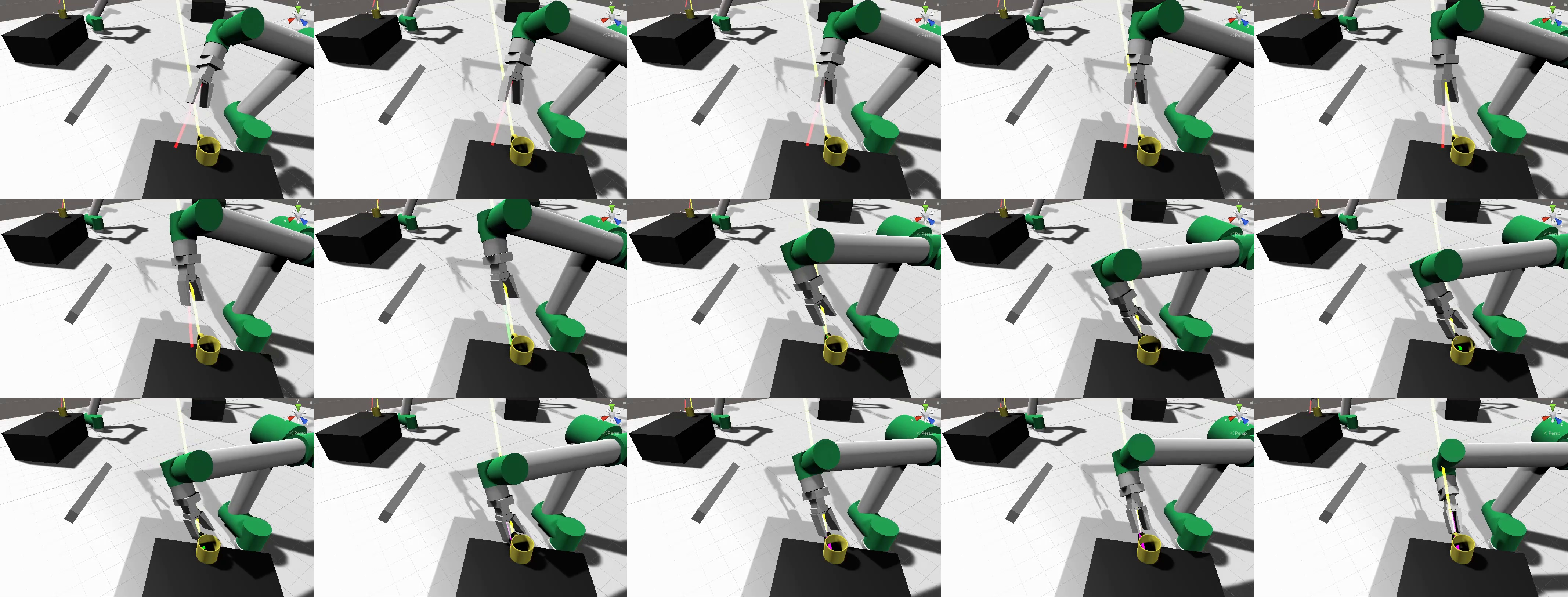}
    \caption{Trained motion}
    \label{fig_learned_motion}
\end{figure*}

At the start of the learning flow shown in Fig. \ref{fig-task-cascade}, imitation learning for Task 1 using GAIL is conducted.
Figure \ref{task1-reward} shows the average cumulative rewards per 1 episode for each 20,000 steps for Task 1.
After 200,000 steps, the algorithm is shifted to reinforcement learning in addition to the GAIL reward.
1.86 million steps later, the reward design was changed to give a small negative reward per step in order to optimize the motion of Task 1.
After about 2.66 million steps, the average cumulative reward and the average number of steps per episode no longer significantly vary (Fig. \ref{tasks-step}) and therefore, learning Task 1 was stopped and shifted to learning Task 2.

The same learning pattern is performed for learning Tasks 2 and 3, with the obvious differences of when (number of steps later) the learning algorithm and reward design were changed as well as the apparent convergence in the average cumulative reward per episode (Fig. \ref{task2-reward},  \ref{task3-reward}).

Finally, in order to optimize the motion throughout the entire task, the reward design was changed so that the cumulative rewards for the next task was added to the reward for the previous task.
The average number of steps per episode no longer decreases after about 500,000 steps from the start of the optimization of the entire task.
The total number of steps required for the overall learning was about 4.6 million steps with the final average number of steps per episode to about 80.

The learned motion is shown in Fig. \ref{fig_learned_motion}.
From our experiments, the imitation of the motion from the demonstration data starts to be learned after about 100,000 to 200,000 steps.
Moreover, we showed that if we try to learn the target motion only by reinforcement learning, the learning may take a long time, or the learning may not be achieved at all.
Therefore, using imitation learning in the initial stage of learning is very effective.
In the demonstration data, it took at least 200 steps to complete 1 episode.
Compared with the demonstration data, the number of steps required for 1 episode was significantly reduced to about 80.

\section{Conclusion}

\subsection{Contribution}
We developed a framework for learning motions based on imitation and reinforcement learning with 
an interface that allows humans to intuitively control the robot in VR space. 
By dividing the grasping motion into tasks, we made it easy for humans to control the robot and also simplify the reward design for reinforcement learning.
Through the VR interface, we were able to collect demonstration data which we used in imitation learning for individual tasks. We also achieved learning of the grasping motion as a whole using reinforcement learning and was able to reduce the number of steps required for the entire task through the combined reinforcement and imitation learning.
For the reward design of reinforcement learning, we further prioritized learning of grasping by changing the reward adaptively according to the learning progress.

\subsection{Future Work}
For practical purposes, we should assume that the robot body is movable. 
In this case, it is necessary to consider moving the robot body and the camera so that it can grasp an object easily.
In this experiment, only one person, the author, operated the robot to get the demonstration data.
If more than one person acquires the demonstration data, there is a possibility that imitation learning will not proceed because the operator's habits will be expressed in the demonstration, which can conflict with the demonstrations of the other operators.
A method for efficiently getting demonstration data by multiple people should be considered.




\begin{thebibliography}{99} 
    \bibitem{estimatemass}  T. Standley, O. Sener, D. Chen, S. Savarese, `` image2mass: Estimating the Mass of an Object from Its Image,'' 1st Annual Conference on Robot Learning, PMLR, pp. 324-333, 2017.
    \bibitem{cornel} Y. Jiang, S. Moseson, and A. Saxena, ``Efficient grasping from RGBD images: Learning using a new rectangle representation,''  International Conference on Robotics and Automation (ICRA), pp. 3304-3311, 2011.
    \bibitem{dex} Q. Lu, K. Chenna, B. Sundaralingam, and T. Hermans, ``Planning multi-fingered grasps as probabilistic inference in a learned deep network,'' International Symposium on Robotics Research (ISRR), 2017.
    \bibitem{soft} C. Choi, W. Schwarting, J. DelPreto, and D. Rus, ``Learning object grasping for soft robot hands,'' IEEE Robotics and Automation Letters, 3(3):2370–2377, 2018.
    \bibitem{imitate1} B. Akgun, M. Cakmak, K. Jiang, and A. L. Thomaz, “Keyframe-based learning from demonstration,” International Journal of Social Robotics, vol. 4, no. 4, pp. 343–355, 2012.
    \bibitem{imitate2} J. Schulman, J. Ho, C. Lee, and P. Abbeel, “Learning from demonstrations through the use of non-rigid registration,” 16th International Symposium on Robotics Research (ISRR), 2013.
    \bibitem{vinci} M. Talamini, K. Campbell, and C. Stanfield, “Robotic gastrointestinal surgery: early experience and system description,” Journal of laparoendoscopic \& advanced surgical techniques, vol. 12, no. 4, pp. 225–232, 2002.
    \bibitem{2dsota} S. Kumra, S. Joshi and F. Sahin, “Antipodal Robotic Grasping using Generative Residual Convolutional Neural Network,” IEEE/RSJ International Conference on Intelligent Robots and Systems (IROS), 2020.
    \bibitem{dexnet} J. Mahler, J. Liang, S. Niyaz, M. Laskey, R. Doan, X. Liu, J. A. Ojea, and K. Goldberg, ``Dex-net 2.0: Deep learning to plan robust grasps with synthetic point clouds and analytic grasp metrics,'' Robotics: Science and Systems (RSS), 2017.
    \bibitem{deep} I. Lenz, H. Lee, and A. Saxena, ``Deep learning for detecting robotic grasps,'' The International Journal of Robotics Research, Vol 34, Issue 4-5, 2015.
    \bibitem{pointnet} C. R. Qi, H. Su, K. Mo, and L. J. Guibas, ``Pointnet: Deep learning on point sets for 3d classification and segmentation,'' IEEE Conference on Computer Vision and Pattern Recognition (CVPR), pp. 77-85, 2017.
    \bibitem{pointnetgpd} H. Liang, X. Ma, S. Li, M. \"{G}orner, S. Tang, B. Fang, F. Sun, and J. Zhang. ``PointNetGPD: Detecting grasp configurations from point sets,'' IEEE International Conference on Robotics and Automation (ICRA), pp. 3629-3635, 2019.
    \bibitem{graspinginteraction} X. Yan, J. Hsu, M. Khansari, Y. Bai, A. Pathak, A. Gupta, J. Davidson, and H. Lee, ``Learning 6-dof grasping interaction via deep geometry-aware 3d representations,'' IEEE International Conference on Robotics and Automation (ICRA), pp. 3766-3773, 2018.
    \bibitem{vr1} C. Stanton, A. Bogdanovych, and E. Ratanasena, “Teleoperation of a humanoid robot using full-body motion capture, example movements, and machine learning,” in Proc. Australasian Conference on Robotics and Automation, 2012.
    \bibitem{vr2} L. Fritsche, F. Unverzag, J. Peters, and R. Calandra, “First-person teleoperation of a humanoid robot,” 2015 IEEE-RAS 15th International Conference on Humanoid Robots (Humanoids), pp. 997-1002, 2015.
    \bibitem{vr3} J. I. Lipton, A. J. Fay, and D. Rus, “Baxter’s homunculus: Virtual
    reality spaces for teleoperation in manufacturing,” IEEE Robotics and
    Automation Letters, vol. 3, no. 1, pp. 179–186, 2018.
    \bibitem{flex} M. Macklin, M. Muller, N. Chentanez, and ¨
    T. Kim, ``Unified particle physics for real-time applications,'' ACM Transactions on Graphics (TOG), 33(4):153, 2014.
    \bibitem{learning6dof} A. Mousavian, C. Eppner, and D. Fox, ``6-dof graspnet: Variational grasp generation for object manipulation,'' International Conference on Computer Vision (ICCV), 2019. 
    \bibitem{google} S. Levine, P. Pastor, A. Krizhevsky, and D. Quillen, ``Learning hand-eye coordination for robotic grasping with deep learning and large-scale data collection,'' The International Journal of Robotics Research, Volume: 37 issue: 4-5, page(s): 421-436, 2018.
    \bibitem{supersizing} L. Pinto and A. Gupta, ``Supersizing self-supervision: Learning to grasp from 50k tries and 700 robot hours,'' IEEE International Conference on Robotics and Automation (ICRA), pages 3406–3413, 2016.
    \bibitem{nothuman} N. Heess, S. Sriram, J. Lemmon, J. Merel, G. Wayne, Y. Tassa, T. Erez, Z. Wang, A. Eslami, M. Riedmiller, et al. “Emergence of Locomotion Behaviours in Rich Environments,” In: arXiv preprint arXiv:1707.02286, 2017.
    \bibitem{fish} J. S. Dyrstad, E. Ruud Øye, A. Stahl, J. Reidar Mathiassen, ``Teaching a Robot to Grasp Real Fish by Imitation Learning from a Human Supervisor in Virtual Reality,'' International Conference on Intelligent Robots and Systems (IROS), pp. 7185-7192, 2018.
    \bibitem{imitation} T. Zhang, Z. McCarthy, O. Jow, D. Lee, K. Goldberg, and
    P. Abbeel. "Deep imitation learning for complex manipulation
    tasks from virtual reality teleoperation," International Conference on Robotics and Automation (ICRA), pp. 5628-5635, 2018.
    \bibitem{bc} S. Ross, G. J. Gordon, and D. Bagnell, “A reduction of imitation learning and structured prediction to no-regret online learning.” Fourteenth International Conference on Artificial Intelligence and Statistics, JMLR Workshop and Conference Proceedings 15:627-635, 2011. 
    \bibitem{irl} A. Y. Ng, S. J. Russell, et al., “Algorithms for inverse reinforcement learning.” 7th International Conference on Machine Learning, pp. 663–670, 2000.
    \bibitem{gail}  J. Ho and S. Ermon, “Generative adversarial imitation learning,” 30th Conference on Neural Information Processing Systems (NIPS), pp. 4565–4573, 2016.
    \bibitem{gae} J. Schulman, P. Moritz, S. Levine, M. I. Jordan, and P. Abbeel, "High-dimensional continuous control using generalized advantage estimation," 4th International Conference on Learning Representations (ICLR), 2016. 

    \bibitem{lstm} S. Hochreiter, and J. Schmidhuber, "Long short-term memory," Neural Computation, 9(8), 1735–1780, 1997.
    \bibitem{ppo} J. Schulman, F. Wolski, P. Dhariwal, A. Radford, and O. Klimov. "Proximal policy optimization algorithms," arXiv preprint arXiv:1707.06347, 2017.


    \end{thebibliography}
\end{document}